%% file: sample-sigconf.tex
\documentclass[sigconf]{acmart}
\acmSubmissionID{2163}

\AtBeginDocument{%
  }

\copyrightyear{2024}
\acmYear{2024}
\setcopyright{rightsretained}
\acmConference[CIKM '24] {Proceedings of the 33rd ACM International Conference on Information and Knowledge Management}{October 21--25, 2024}{Boise, ID, USA.}
\acmBooktitle{Proceedings of the 33rd ACM International Conference on Information and Knowledge Management (CIKM '24), October 21--25, 2024, Boise, ID, USA}
\acmISBN{979-8-4007-0436-9/24/10}
\acmDOI{10.1145/3627673.3679838}

\input{texts/commands}

\usepackage{graphicx,tabularx}
\usepackage{multirow}
\usepackage{multicol}
\usepackage{subcaption}
\usepackage{balance}

\begin{document}

\title{MVMR: A New Framework for Evaluating Faithfulness of Video Moment Retrieval against Multiple Distractors}

\author{Nakyeong Yang}
\affiliation{
  \institution{Seoul National University}
  \city{Seoul}
  \country{Korea}
}
\email{yny0506@snu.ac.kr}

\author{Minsung Kim}
\affiliation{
  \institution{Seoul National University}
  \city{Seoul}
  \country{Korea}
}
\email{kms0805@snu.ac.kr}

\author{Seunghyun Yoon}
\affiliation{
  \institution{Adobe Research}
  \city{San Jose}
  \country{USA}
}
\email{syoon@adobe.com}

\author{Joongbo Shin}
\affiliation{
 \institution{LG AI Research}
 \city{Seoul}
 \country{Korea}
}
\email{jb.shin@lgresearch.ai}

\author{Kyomin Jung}
\affiliation{
  \institution{Seoul National University}
  \city{Seoul}
  \country{Korea}
}
\email{kjung@snu.ac.kr}



\input{texts/abstract}





\maketitle

\section{Introduction}
\input{texts/intro.tex}

\section{Video Moment Retrieval}
\input{texts/relworks.tex}

\section{MVMR Problem Definition}
\input{texts/problem_def.tex}

\section{MVMR Dataset Construction}
\label{ssec:mvmr_dataset_construction}
\input{texts/method_dataset.tex}

\section{MVMR Dataset Analysis}
\label{ssec:mvmr_dataset_analysis}
\input{texts/method_dataset_analysis.tex}

\section{Methods: CroCs}
\input{texts/method_train.tex}

\section{Experiments}
\input{texts/experiments.tex}

\section{Conclusion}
\input{texts/conclusion.tex}

\input{texts/ack.tex}

\appendix
\input{texts/appendix.tex}

\bibliographystyle{ACM-Reference-Format}
\balance
\bibliography{sample-base}

\end{document}

%% file: texts/commands.tex
\newcommand{\prevtask}{VMR}
\newcommand{\prevtaskvcmr}{VCMR}
\newcommand{\ourtask}{MVMR}

%% file: texts/abstract.tex
\begin{abstract} 
With the explosion of multimedia content, video moment retrieval (\prevtask), which aims to detect a video moment that matches a given text query from a video, has been studied intensively as a critical problem.
However, the existing \prevtask~framework evaluates video moment retrieval performance, assuming that a video is given, which may not reveal whether the models exhibit overconfidence in the falsely given video.
In this paper, we propose the \textbf{\ourtask} (\textbf{M}assive \textbf{V}ideos \textbf{M}oment \textbf{R}etrieval for Faithfulness Evaluation) task that aims to retrieve video moments within a massive video set, including multiple distractors, to evaluate the faithfulness of \prevtask~models.
For this task, we suggest an automated massive video pool construction framework to categorize negative (distractors) and positive (false-negative) video sets using textual and visual semantic distance verification methods. We extend existing \prevtask~datasets using these methods and newly construct three practical \ourtask~datasets.
To solve the task, we further propose a strong informative sample-weighted learning method, CroCs, which employs two contrastive learning mechanisms: (1) weakly-supervised potential negative learning and (2) cross-directional hard-negative learning. 
Experimental results on the \ourtask~datasets reveal that existing \prevtask~models are easily distracted by the misinformation (distractors), whereas our model shows significantly robust performance, demonstrating that CroCs is essential to distinguishing positive moments against distractors. Our code and datasets are publicly available: \href{https://github.com/yny0506/Massive-Videos-Moment-Retrieval}{link}.

\input{file_sources/fig1.tex}
\end{abstract}

%% file: file_sources/fig1.tex
\begin{figure}[b]
\centering
\includegraphics[width=0.9\linewidth]{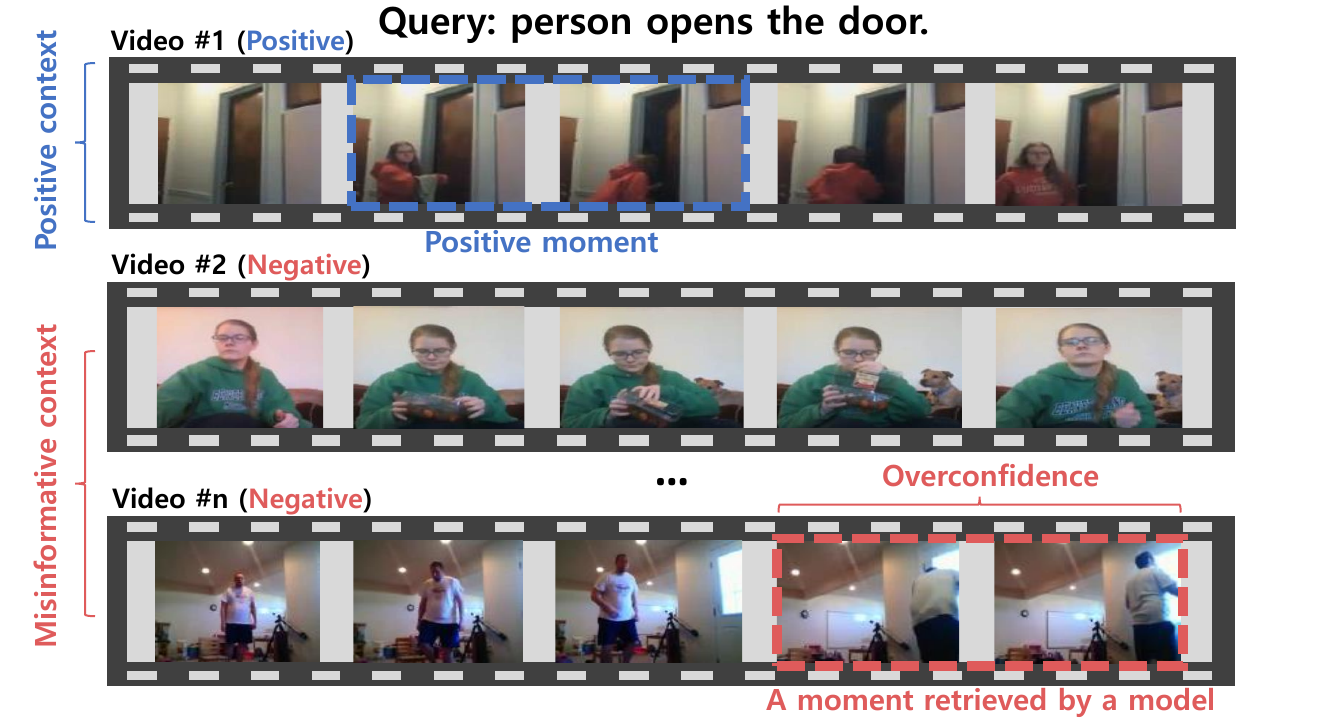}
\vspace{-0.2cm}
\caption{Faithfulness Issue in Video Moment Retrieval. If a VMR model is trained without considering overconfidence in a given context (video), it may undergo a Type 1 error since it prioritizes selecting a video moment from the falsely given video (distractor).}
\vspace{-0.4cm}
\label{fig:fig1}
\end{figure}


%% file: texts/intro.tex
Enabled by the increased accessibility of video-sharing platforms along with the advancement of networking and storage technology, a vast amount of new content becomes available on the web daily. 
With this huge number of contents, Video Moment Retrieval (\prevtask) task, searching for a video moment suitable for a given query, has emerged as an essential problem \citep{anne2017localizing, gao2017tall}.

However, this line of research evaluates retrieval models assuming that a context (video) is given, it is limited in revealing the capability of distinguishing misinformation.
Figure~\ref{fig:fig1} exemplifies the problem of faithfulness for misinformation demonstrated by a \prevtask~model.
This overconfidence issue has been studied recently in the natural language community as a challenging task \citep{chen2022towards, shi2023trusting}, but it has not been studied sufficiently in the \prevtask~field.
This problem also arises in a practical video moment retrieval scenario, Video Corpus Moment Retrieval (VCMR), which aims to search for a relevant video moment within a pool of multiple videos \citep{escorcia2019temporal, lei2020tvr, ma2022interactive, jung2022modal}.
The VCMR framework evaluates video moment retrieval performance using a sequential pipeline that integrates video retrieval with video moment retrieval.
Therefore, it also diverges from evaluating the overconfidence of a \prevtask~model, as the evaluated retrieval performance may predominantly depend on video retrieval capability.

\input{file_sources/fig2.tex}

To tackle this issue, we propose an \textbf{\ourtask} (\textbf{M}assive \textbf{V}ideos \textbf{M}oment \textbf{R}etrieval for Faithfulness Evaluation) task, which expands the search coverage to a massive video set that can include any number of positive and negative videos for evaluating the faithfulness of VMR models.
Specifically, given a text query $q$ with positive videos $v^{+}_{1}, ..., v^{+}_{k}$ and negative videos $v^{-}_{1}, ..., v^{-}_{n-k}$, the \ourtask~task aims to retrieve temporal moments in the positive videos that match the query from a massive video set $V^{+,-}_{q}=\{v^{+}_{1},...,v^{+}_{k}, v^{-}_{1}, ..., v^{-}_{n-k}\}$ without the interference of any video retrieval model.

For the \ourtask~task, we further construct practical \ourtask~benchmarks by extending publicly available \prevtask~datasets.
Specifically, we categorize positive and negative videos for a specific query from the whole video set of a \prevtask~dataset to construct a massive video set.
The most critical challenge in converting a \prevtask~dataset into an \ourtask~dataset is accurately categorizing positive and negative videos for each query.
However, addressing this issue by manually annotating a large number of videos for each query to determine whether they are positive or negative is exceedingly labor-intensive.
Therefore, we propose a new automated framework to construct \ourtask~datasets by employing textual and visual semantic distance verification methods to categorize positive and negative videos for each query, as shown in Figure~\ref{fig:fig2}.
Specifically, we examine the distance between a target query and all videos in a \prevtask~dataset using state-of-the-art text embedding model \cite{gao2021simcse} and video-captioning evaluation model \cite{shi2022emscore} to define the positive and negative massive video sets.
Consequently, we construct three \ourtask~datasets, and we confirm that the introduced datasets contain only 1.5\%, 0.2\%, and 3.5\% falsely defined videos from human evaluation, showing that our approach successfully constructs practical benchmarks.
On these benchmarks, we evaluate strong baseline \prevtask~models, including state-of-the-art models, and demonstrate that significant overconfidence issues arise in the new scenario (\ourtask).
For example, the state-of-the-art model, MMN, shows an average 35.9\% degradation in the performance (R1@0.5 and R5@0.5) for all \ourtask~datasets, revealing the difficulty of the \ourtask~task.

In addition, to enhance the faithfulness of \prevtask~models, we introduce a novel informative sample-weighted learning, CroCs, which stands for \textbf{Cro}ss-directional informative sample-weighted \textbf{C}ontra\textbf{s}tive Learning.
Specifically, CroCs adopts two training mechanisms: (1) weakly-supervised potential negative learning and (2) cross-directional hard-negative learning.
The weakly-supervised potential negative learning assesses the distance between queries to exclude false-negatives.
The cross-directional hard-negative learning measures the relevance score between queries and moments predicted by a model to exclude easy negative samples.

We demonstrate that our CroCs outperforms strong baselines significantly in the \ourtask~setting, and this result proves that informative sample-weighted constrastive learning successfully mitigates the misinformation overconfidence of a \prevtask~model.
Moreover, we show that CroCs is also effective in the realistic scenario by adopting the widely-used pipeline, which includes a video retrieval model. In summary, this work makes the following contributions:
\begin{itemize}
\item We propose a new task called \ourtask, and construct three \ourtask~datasets to evaluate the faithfulness of \prevtask~models.

\item We propose a novel automated dataset construction framework, the first to consider how to construct a trustful massive video retrieval pool via semantic distance verification.

\item We demonstrate that existing \prevtask~models are significantly vulnerable to misinformation; thus, we introduce a novel informative sample-weighted learning method, CroCs, to enhance the faithfulness of \prevtask~models against misinformation.

\end{itemize}

\input{file_sources/fig_dataset_qualitative_analysis}

%% file: file_sources/fig2.tex
\begin{figure}[t]
\centering
\includegraphics[width=0.9\linewidth]{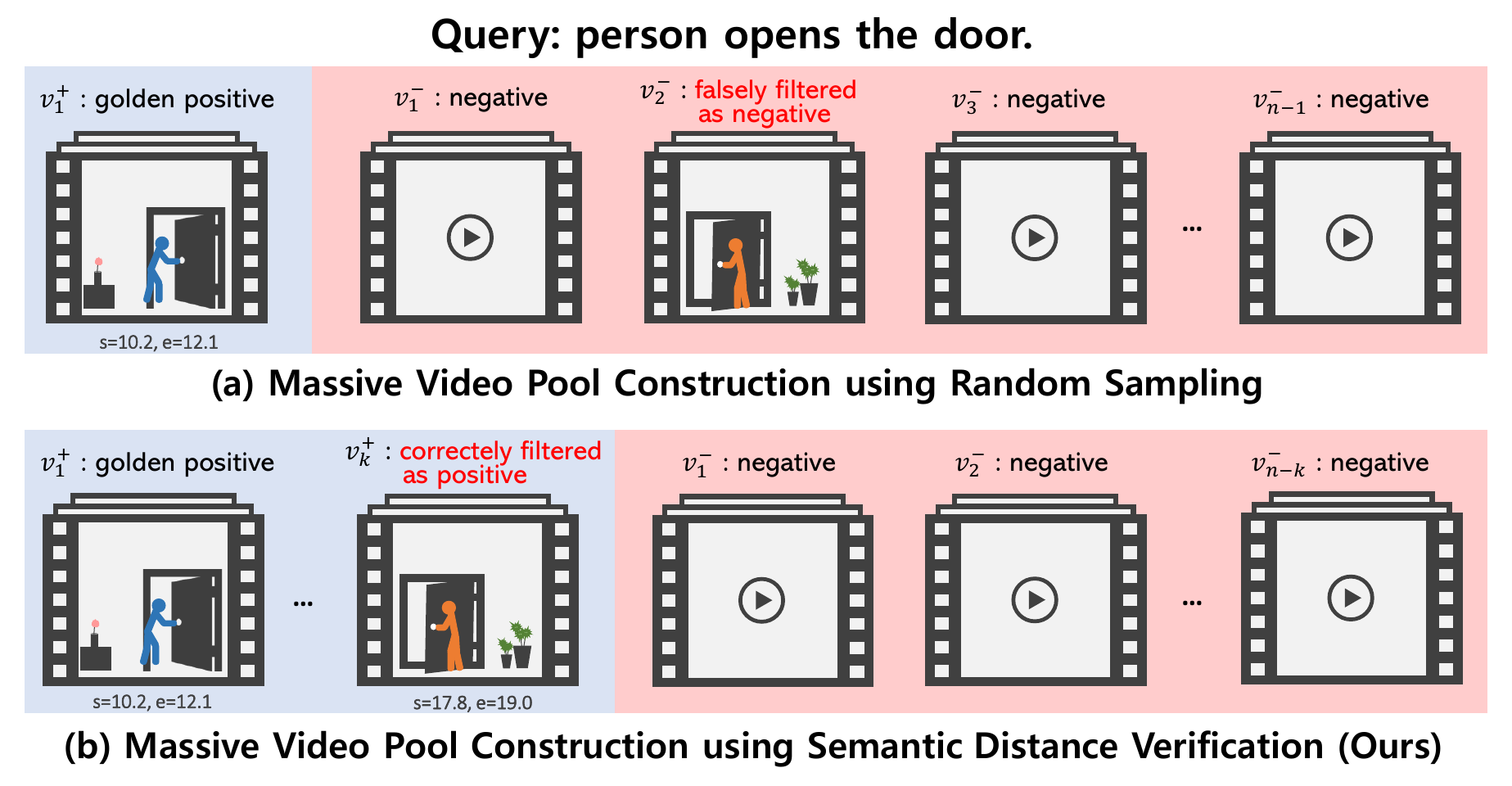}
\vspace{-0.2cm}
\caption{Massive Video Pool Construction.
From our experiment on the TACoS dataset, if a massive video set is constructed using random sampling, then over 40\% of queries include at least one false-negative video, showing the risk of random sampling.
Our method constructs a massive video set by filtering trustful positive and negative sets using semantic distance verification methods, considering the possibility of false-negatives. $v^{+}_{i}$ and $v^{-}_{j}$ mean a positive and a negative video, respectively.
}
\vspace{-0.5cm}
\label{fig:fig2}
\end{figure}

%% file: file_sources/fig_dataset_qualitative_analysis.tex
\begin{figure*}[hbtp]
\newcommand\x{12}
\newcommand\p{1.0}

\begin{subfigure}[t]{\p\textwidth}
    \centering
    \includegraphics[width=\textwidth]{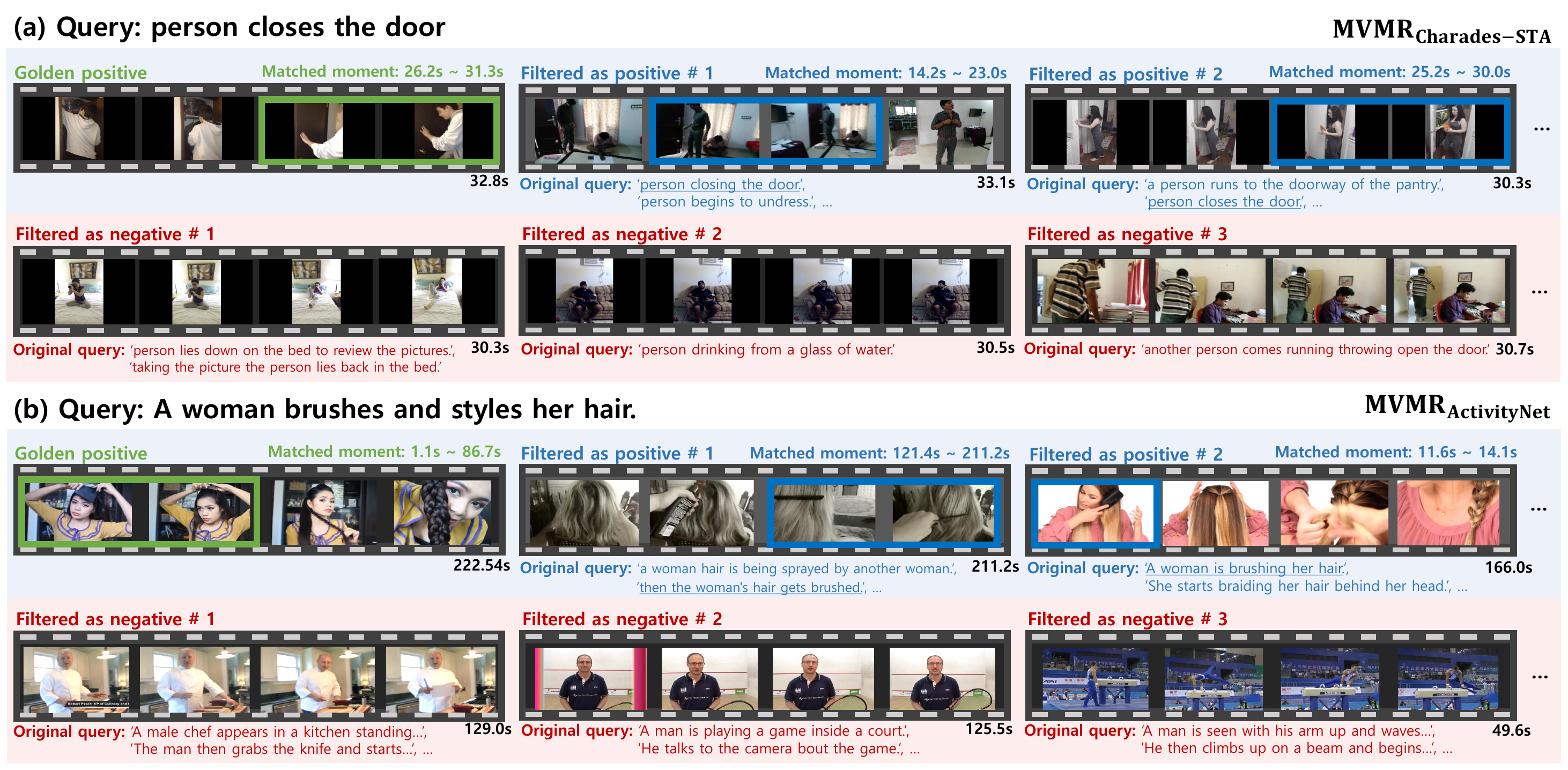}
\end{subfigure}
\vspace{-0.5cm}

\caption{\textbf{Examples of constructed MVMR datasets.} We visualize positive and negative video sets for a query of the constructed three \ourtask~datasets. 
A green solid box means a golden positive moment, and blue solid boxes show moments assigned to videos classified as positive. The underlined queries mean the most similar query described in Section \ref{ssec:semantic_check}~(\textit{max aggregation}).}
\label{fig:fig_qualitative}
\vspace{-0.2cm}
\end{figure*}

%% file: texts/relworks.tex
Video Moment Retrieval (\prevtask), which aims to detect a specific moment in a video that matches a given text query, is one of the most challenging problems in the video-language multi-modal domain.
Therefore, the \prevtask~task has been explored extensively using numerous effective methods ~\citep{zhang2020learning, zhang2020span, nan2021interventional, gao2021fast, liu2021context, wang2022negative, li2023g2l}.
Some novel approaches have regarded the problem as question-answering (QA) tasks \citep{seo2018bidirectional, joshi2020spanbert}, by leveraging a QA model to encode a multi-modal representation and then predicting the frames that correspond to the start and end of the moment that match a query \citep{ghosh2019excl, zhang2020span}.
However, \prevtask~studies operate within an impractical framework, focusing on searching moments within a given video.
Consequently, their settings are limited in retrieving relevant information across all media repositories and have not evaluated the faithfulness of VMR models on misinformation.

Existing studies \citep{escorcia2019temporal, lei2020tvr, ma2022interactive, jung2022modal} have extended the \prevtask~to search on a multiple video pool, naming the task as the Video Corpus Moment Retrieval (\prevtaskvcmr), which focuses on retrieving a moment from multiple videos for a given query.
The VCMR task considers the most practical usage scenario of VMR. However, it is still limited by its inability to evaluate the faithfulness of VMR models in a misinformative context.
Existing VCMR studies have utilized a framework evaluating moment retrieval performance by integrating video retrieval with VMR; thus, the evaluated performance in the VCMR setting may overwhelmingly depend on video retrieval capability.
In addition, the VCMR framework simply extends the \prevtask~task by categorizing only a paired positive video moment for a target query as positive while labeling all unpaired videos as negative videos \citep{10095182, zhang2021video}.
This approach deviates from the practical scenario of the video moment search since multiple positive videos can be present in whole media collections for a target query.
Moreover, existing VCMR studies have randomly selected negative samples from all unpaired videos; thus, they have overlooked the possibility of false-negatives when constructing the negative video set for a query, which raises significant trustfulness concerns.

Some studies \citep{Yoon2022SelectiveQD, hao2022shufflevideos, 10095182} have attempted to address unusual behaviors of VMR models, such as bias or spurious correlation, which arise during the training for the \prevtask~and VCMR tasks.
\citet{Yoon2022SelectiveQD} has categorized good and bad biases to train a model using only a good bias.
\citet{hao2022shufflevideos} has mitigated a temporal bias by training a model with temporally shuffled moments.
However, existing studies have never addressed the overconfidence issue in a misinformative context; thus, it remains a challenging task to solve.

%% file: texts/problem_def.tex

The MVMR task evaluates VMR models under a massive video pool with $k$ positive videos and $n-k$ negative videos for each query $q$.
Given a positive video set $V^{+}_{q} = \{v^{+}_{1}, ..., v^{+}_{k}\}$ and a negative video set $V^{-}_{q} = \{v^{-}_{1}, ..., v^{-}_{n-k}\}$, the \ourtask~task aims to search for a moment $m = (x_{s}, x_{e})$ of a specific video $v$ that matches the query from massive video set $V^{+,-}_{q}=\{v^{+}_{1},...,v^{+}_{k}, v^{-}_{1}, ..., v^{-}_{n-k}\}$, where $x_{s}$ and $x_{e}$ mean start and end points of the moment, respectively.
In contrast to the VCMR task, we only utilize a VMR model $\mathcal{P}$ by inputting a target query $q$ and each video $v$ in $V^{+,-}_{q}$ to compute the matching score $p = \mathcal{P}(m|q, v)$ for each moment $m$.
We finally determine the optimal moment $m^{*} = (x^{*}_{s}, x^{*}_{e})$ by comparing the matching score of all moments in $V^{+,-}_{q}$ as $m^{*} = argmax_{m} \mathcal{P}(m|q, v)$.

\input{file_sources/table_dataset_statics}

%% file: file_sources/table_dataset_statics.tex
\begin{table*}[th]
\centering
\resizebox{\textwidth}{!}{
\begin{tabular}{@{}lcccccc@{}}
\toprule
\hspace{2.5em}\multirow{2}{*}{\textbf{Dataset}} & \hspace{1.5em}\multirow{2}{*}{\textbf{\# Queries / \# Videos}}\hspace{2.8em} & \hspace{3.8em}\textbf{Avg Len (sec)}\hspace{3.8em} & \hspace{3.8em}\textbf{Avg \# Moments}\hspace{3.8em} & \hspace{3.8em}\textbf{Avg}\hspace{3.8em} & \hspace{3.5em}\textbf{\# Max Positive /}\hspace{3.5em} \\

& & \textbf{Moment / Video} & \textbf{per Query} & \textbf{Query Len} & \textbf{\# Retrieval Pool} \\
\midrule

$\text{Charades-STA}$ & 3720 / 1334 & 7.83 / 29.48 & 1 & 6.24 & 1 / 1\\
$\text{ActivityNet}$ & 17031 / 4885 & 40.25 / 118.20 & 1 & 12.02 & 1 / 1\\
$\text{TACoS}$ & 4083 / 25 & 31.87 / 367.15 & 1 & 8.53 & 1 / 1\\
\midrule
$\text{\ourtask}_\text{Charades-STA}$ & 3716 / 1334 & 7.83 / 29.48 & 3.07 & 6.23 & 5 / 50\\
$\text{\ourtask}_\text{ActivityNet}$ & 16941 / 4885 & 40.32 / 118.20 & 1.11 & 12.03 & 5 / 50\\
$\text{\ourtask}_\text{TACoS}$ & 2055 / 25 & 28.55 / 367.15 & 2.24 & 7.31 & 5 / 5\\
\bottomrule
\end{tabular}}
\caption{\textbf{Summary of datasets.} Remarkably, the queries of $\text{\ourtask}_\text{Charades-STA}$ and $\text{\ourtask}_\text{TACoS}$ generally include multiple positive moments, and it empirically proves that a massive video set construction using random sampling is inadequate to construct the \ourtask~dataset since it is exposed to the risk of false negatives.}
\vspace{-0.7cm}
\label{tab:dataset_stats}
\end{table*}

%% file: texts/method_dataset.tex
Our MVMR setting automatically reconstructs a \prevtask~dataset by expanding it to associate each query with multiple positive and negative moments.
Each \prevtask~dataset consists of multiple videos with some query-moment pairs, and our method carefully categorizes the positive and negative sets using two semantic distance verification techniques: Query Distance-based Verification and Query-video Distance-based Verification.
In the following, we detail our process for constructing the \ourtask~benchmark and present the analyses of three constructed \ourtask~datasets.

\subsection{Semantic Distance Verification}
\label{ssec:semantic_check}
This section describes the verification process of the distance between a target query $q$ and a video $v$ using semantic text embedding and video captioning evaluation models.

\paragraph{\textbf{Query Distance-based Verification.}}
\prevtask~datasets consist of video set $V=\{v_{1},...,v_{l}\}$ and paired query set $Q_{v} = \{q_{1}, ..., q_{m}\}$ for each video $v$, where $l$ and $m$ is the number of videos and text queries included in a dataset, respectively.
Since each text query semantically describes the paired moment, we use queries to obtain semantic information about the paired video.
For example, suppose that a video $v$ is paired with a query $q$, \textit{"The person gets out a knife"}, in a moment $(x_{s}, x_{e})$. 
Since a target query $\hat{q}$, \textit{"The person takes out a knife"}, is highly similar to the query $q$, $\hat{q}$ is associated with the moment $(x_{s}, x_{e})$ of the video $v$.
We use a semantic text embedding model, SimCSE \cite{gao2021simcse}, to distinguish positive and negative videos for each query.
Specifically, we first obtain embeddings $e_{\hat{q}}$ and $\{e_{q_{1}}, ..., e_{q_{m}}\}$ for the target text query $\hat{q}$ and all video text queries $\{q_{1}, ..., q_{m}\}$ using SimCSE, respectively.
Then, we calculate cosine similarity $s_{te}(\hat{q}, q_{i})$ between $e_{\hat{q}}$ and $e_{q_{i}}$ to quantify the semantic matching, and select the maximum similarity among $\{s_{te}(\hat{q}, q_{1}),...,s_{te}(\hat{q}, q_{m})\}$ to use it as the query-video matching score $s_{te}(\hat{q}, v)$ of the target query $\hat{q}$ and the video $v$.
We select \textit{max aggregation} to obtain $s_{te}(\hat{q}, v)$ since the video should be regarded as a positive video, even if only one similar query to a target query exists in a video.

\paragraph{\textbf{Query-video Distance-based Verification.}}
Paired queries may describe the video limitedly since they may not cover all its context.
We examine a visual semantic matching between a query and a video to categorize positive and negative videos.
We use EMScore \cite{shi2022emscore}, a video captioning evaluation model, to quantify the visual similarity $s_{tv}(q,v)$, used as a complementary filtering method on videos primarily selected through the query similarity $s_{te}(q,v)$.

\subsection{Categorizing Positive and Negative sets}
\label{ssec:filtering}
This section describes constructing positive and negative video sets using the calculated query-video similarities, $s_{te}$ and $s_{tv}$. 
Positive candidates are identified by primarily excluding videos with $s_{te}(q, v) < t_{te}^{+}$, where $t_{te}^{+}$ is a hyper-parameter. We further exclude videos that have lower similarity than the mean $s_{tv}$ of all golden positive query-video pairs $(q, v^{*})$ for all queries.
We construct final \ourtask~positives by randomly sampling $k$ videos from this categorized set of positive candidates.
Negative candidates are identified by excluding videos with $s_{te}(q, v) > t_{te}^{-}$, where $t_{te}^{-}$ is also a hyper-parameter. We additionally exclude videos that have higher similarity than the mean $s_{tv}$ of query-video pairs $(q, v^{-})$, which are primarily categorized as negatives using $s_{te}$.
Like \ourtask~positives, we construct final \ourtask~negatives by randomly sampling $(n-k)$ videos from this set of negative candidates.
From this process, a total of $n$ video retrieval pool $V^{+,-}_{q}=\{v^{+}_{1},...,v^{+}_{k}, v^{-}_{1}, ..., v^{-}_{n-k}\}$ is obtained for each query.

%% file: texts/method_dataset_analysis.tex
\subsection{\ourtask~Settings.}
This section describes the utilized \prevtask~datasets for our \ourtask~framework and detailed settings to derive \ourtask~positive and negative candidates using SimCSE and EMScore.

\paragraph{\textbf{\prevtask~Datasets.}}
We extend widely-used \prevtask~datasets, Charades-STA \cite{gao2017tall}, ActivityNet \cite{krishna2017dense}, and TACoS \cite{regneri2013grounding}, to construct three \ourtask~benchmarks.
We name the constructed \ourtask~datasets as $\text{\ourtask}_{\textsc{\{\$datasource\}}}$.

\paragraph{\textbf{Filtering Settings.}}
We use the STS16 dataset \cite{agirre2016semeval} to empirically validate varying SimCSE filtering hyper-parameters, $t^{+}_{te}$ and $t^{-}_{te}$.
The dataset provides text pairs and human-annotated similarity scores (five-point Likert scale), and two texts are considered similar when the score exceeds 3.
We analyze the distribution of SimCSE and human annotated scores for all text pairs in the dataset, and select $t^{+}_{te}=0.9$ to derive MVMR positive candidates since it is regarded as the best threshold to distinguish similar queries.
Similarly, we select $t^{-}_{te}=0.5$ to identify \ourtask~negative candidates.
The distribution of SimCSE and human-annotated scores are shown in Appendix~\ref{sec:appendix_analysis_filters}.

\input{file_sources/fig_dist_all}

\subsection{\ourtask~Analysis}
We fixedly set the size of the massive video pool as $n$, containing at most $k$ positive videos for each query.
If the number of \ourtask~positive candidates derived for each query is less than $k$, we include all positive candidates as positive samples.
Also, if the total number of \ourtask~positive and \ourtask~negative candidates for a query is less than $n$, the query is excluded from the final \ourtask~dataset.

\paragraph{\textbf{Datasets Statistic.}}
We determine the number of videos in the $\text{\ourtask}_\text{Charades-STA}$ and the $\text{\ourtask}_\text{ActivityNet}$ retrieval pools as $n=50$, and the maximum number of positive moments per query $k$ is set to $5$.
The former consists of 3,716 queries for 1,334 videos, each containing an average of 3.07 positive moments.
The latter contains 16,941 queries for 4,885 videos, each including an average of 1.11 positive videos.
For the $\text{\ourtask}_\text{TACoS}$ dataset, We set the size of the video pool and the maximum number of positive moments per query as $5$ and $5$, respectively, since the TACoS test set contains only 25 videos totally.
It includes 2,055 queries for 25 videos, each containing an average of 2.24 positive videos. The summary of the utilized \prevtask~datasets and the constructed \ourtask~datasets is shown in Table \ref{tab:dataset_stats}.

\paragraph{\textbf{Dataset Quality.}}
Figure~\ref{fig:fig_qualitative} shows categorized positive and negative examples for a query of the constructed \ourtask$_{\text{Charades-STA}}$ and \ourtask$_{\text{ActivityNets}}$ datasets.
The figure indicates that matched videos to a query are correctly classified as positive.



\paragraph{\textbf{Analysis on Semantic Distance Verification.}}
We visualize the results derived using SimCSE and EMScore to analyze our MVMR datasets qualitatively.
Figure~\ref{fig:fig_all_dist}-(a) displays SimCSE embeddings of all queries in each dataset, illustrating the similarity between all queries and a specific target query, \textit{"person turn a light on."} (left) and \textit{"He is using a push mower to mow the grass."} (right).
The figure of TACoS can also be found in Appendix~\ref{sec:appendix_analysis_filters}.
The figures show that queries exhibit well-defined clusters for Charades-STA and TACoS, revealing the effectiveness of our Query Distance-based filtering.
These results are also supported by Figure~\ref{fig:fig_all_dist}-(b) SimCSE score histogram, showing that Charades-STA and TACoS have many similar queries.
Figure~\ref{fig:fig_all_dist}-(c) illustrates histograms of EMScore similarity between queries and videos. The red and blue histograms correspond to the EMScore distribution of positive and negative samples, respectively. In the case of ActivityNet, EMScore effectively distinguishes positives and negatives. This success is attributed to that ActivityNet includes detailed queries and videos with diverse features.
According to the findings presented in this section, the Semantic Distance Verification method's efficacy varies depending on the dataset's characteristics. Consequently, a combination of two filtering methods (i.g., SimCSE and EMScore) should be employed in a complementary manner to ensure the construction of a trustful \ourtask~dataset.

\paragraph{\textbf{Human Evaluation.}}
We recruit crowd workers fluent in English through the university’s online community.
We ask them to examine 100 queries manually and their paired videos (by our method) and answer whether they are correctly labeled.
Overall, we analyze 4,900, 4,900, and 400 videos for Charades-STA, ActivityNet, and TACoS, respectively, since each \ourtask~dataset consists of 49, 49, and 4 newly added videos (either positive or negative samples). The original videos (i.e., golden positive samples) are excluded from this investigation.
Based on human evaluation, we confirm that the newly introduced dataset contains only 1.5\%, 0.2\%, and 3.5\% falsely categorized videos for the three datasets, respectively.
TACoS exhibits a relatively higher query-relevant video ratio than the other datasets since its test set includes only 25 videos and consists of significantly similar cooking activities within the kitchen.

%% file: file_sources/fig_dist_all.tex
\begin{figure*}[!h]
\newcommand\x{12}
\newcommand\p{1.0}

\begin{subfigure}[t]{\textwidth}
    \centering
    \includegraphics[width=\p\textwidth]{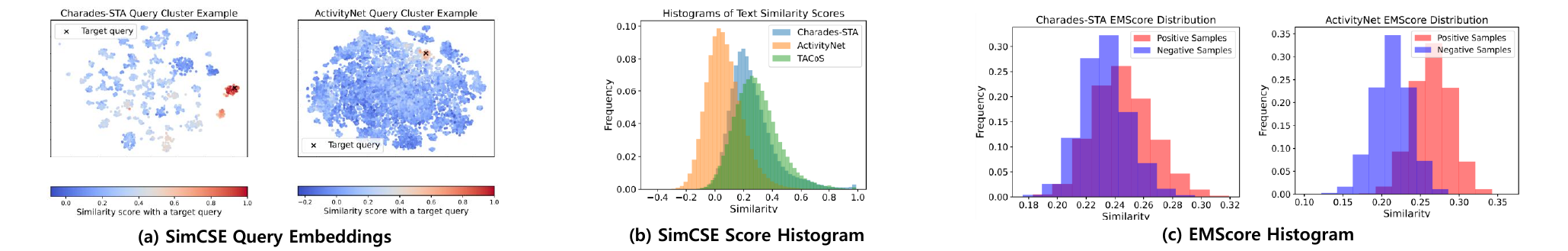}
\end{subfigure}
\vspace{-0.7cm}
\caption{\textbf{Qualitative Analysis for Filtering Methods.} We visualize the derived similarity scores of SimCSE and EMScore to verify the constructed MVMR datasets. We use T-SNE to reduce the dimension of each query embedding for displaying each query (dot) of SimCSE Similarity Distribution.}
\label{fig:fig_all_dist}
\end{figure*}

%% file: texts/method_train.tex
\input{file_sources/fig_method}

To solve the \ourtask~task, this section introduces a novel informative sample-weighted learning method, \textbf{CroCs}, which stands for \textbf{Cro}ss-directional informative sample-weighted \textbf{C}ontra\textbf{s}tive Learning.
A key challenge in solving the \ourtask~task is distinguishing the representation of positives from multiple negative distractors.
Therefore, we select MMN \cite{wang2022negative} as the backbone of our model since it is the optimal baseline for the \ourtask~task since it uses contrastive learning to distinguish negative samples from positives effectively.
On the MMN architecture, we apply CroCs in contrastive learning procedures to solve the challenge in the \ourtask~task.
MMN has limitations in neglecting the potential presence of false-negative video moments since it regards all randomly selected in-batch samples as negatives in contrastive learning.
Furthermore, it overlooks the importance gap among negative samples.
Our CroCs solves these two limitations of MMN by adopting two training mechanisms: (1) Weakly-supervised potential negative learning and (2) Cross-directional hard-negative learning.

\subsection{Weakly-supervised Potential Negative Learning}
\label{ssec:false-neg_contrast}

MMN randomly selects negative samples from the whole samples for contrastive learning.
This random sampling process has a significant limitation due to the potential presence of false-negatives since it categorizes all samples, except for the golden one, as negative training samples.
To mitigate this concern, we propose excluding false-negative videos to refine the contrastive learning process. Specifically, we adopt a potential negative filtering method to identify and exclude false-negative videos as follows:

\footnotesize
\begin{equation}
\begin{aligned}
    V^{-}_{tr}(q) = \{v|s_{te}(q, v) < t_{tr}\}, \:\:\ \forall{v} \in V \:\\
\end{aligned}
\label{eq:sampling}
\end{equation}
\normalsize

where $V^{-}_{tr}(q)$ is a negative sample set of a specific query $q$ to use during the contrastive learning procedure. $s_{te}(q,v)$ is the query-video similarity described in Section \ref{ssec:semantic_check}. $t_{tr}$ is a hyper-parameter for the SimCSE score filtering. 
If a specific video $v$ shows a value of $s_{te}(q,v)$ above the threshold $t_{tr}$ for a target query $q$, we classify $v$ as false-negative for $q$. Therefore, we exclude $v$ in negative samples of contrastive learning for $q$.

\subsection{Cross-directional Hard-negative Learning}
\label{ssec:hard-neg_contrast}

Easy or false-negative samples may distract a model from learning proper knowledge during the training procedure.
Inspired by recent progress in using hard-negative sampling for the document retrieval task \cite{zhou2022simans}, we classify negative samples according to the matching score between a target query and negative moments as follows: (1) Negative samples that are clearly irrelevant and have low matching scores should be sampled less frequently (Easy-negative); (2) Negative samples that are highly relevant and have high matching scores should also be sampled less frequently (False-negative); (3) Negative samples that are uncertain and have matching scores similar to true-positives should be sampled more frequently since they provide useful information (Hard-negative).
Based on these criteria, we distinguish informative hard-negative samples and adopt a retraining process for a model pre-trained on a specific dataset, using only hard-negative samples. We define the negative sampling distribution for the moment and query as follows: 

\scriptsize
\begin{equation}
\begin{aligned}
    p'(m|q) \propto \exp{(-(r(q,m)-\overline{r}(q,m^{+}))^{2})}, \:\ \forall m \in M^{-}_{q} \\
    p'(q|m) \propto \exp{(-(r(q,m)-\overline{r}(q^{+},m))^{2})}, \:\ \forall q \in Q^{-}_{m} \:\:\\
\end{aligned}
\label{eq:mqans1}
\end{equation}
\normalsize

\noindent where $r(q, m)$ is a matching score between a query and a moment calculated using a pre-trained video moment retrieval model, and $\overline{r}(q, m^{+})$ and $\overline{r}(q^{+}, m)$ is mean matching scores for all golden positive moments and queries, respectively. $M^{-}_{q}$ and $Q^{-}_{m}$ is potential negative candidates for a query $q$ and a moment $m$, respectively. $p’(m|q)$ means the probability that a moment $m$ is sampled as negative when given a query $q$. 
Specifically, $p’(m|q)$ is computed by the distance between $r(q,m)$ and $\overline{r}(q,m^{+})$; thus, the closer the distance, the more frequently sampled as negative since a closer sample is a hard-negative sample not easy to distinguish by a model.
In other words, negative candidate moments near the average matching score of golden query-moment pairs have a high sampling probability.
Likewise, $p’(q|m)$ means the probability that a query $q$ is sampled as negative when given a moment $m$.
We also reformulate $p’(m|q)$ to $p’(v|q)$ for computational efficiency in a sampling procedure as follows:

\scriptsize
\begin{equation}
\begin{aligned}
    r(q,v) = \max{(\{r(q,m_{1}), ..., r(q,m_{l})\})}, \:\:\ \forall m_{i} \in v \:\:\:\:\:\:\\
    p'(v|q) \propto \exp{(-(r(q,v)-\overline{r}(q,v^{+}))^{2})}, \:\:\ \forall v \in V^{-}_{q} \:\:\:\:\:\:\\
\end{aligned}
\label{eq:vqans}
\end{equation}
\normalsize

$p'(q|v)$ can be formulated similarly by calculating $\overline{r}$ for positive queries about each video.

\subsection{Informative Sample-weighted Mutual Matching Learning}
This section describes an informative sample-weighted contrastive learning method, CroCs.

\paragraph{\textbf{Model architecture.}} We adopt a bi-encoder architecture with a late modality fusion by an inner product in the joint visual-text representational space following the baseline MMN.
By adopting a bi-encoder architecture, we can efficiently search for moments with pre-computed video moment representations when given a user query.
Our whole model architecture is shown in Figure~\ref{fig:method}.
Our model represents each query and video moment using query and video encoders to two joint visual-text spaces (IoU and mutual matching spaces).
From the joint visual-text spaces, we compute the matching score between query and video moment representations and select the best video moment.
The details of the model architecture are described in the appendix~\ref{sec:appendix-rmmn}.

\paragraph{\textbf{Informative Sample-weighted Mutual Matching.}} We select the binary cross entropy and contrastive learning losses and train our model following two steps: (1) the first training step using the two loss functions (binary cross entropy and contrastive learning losses) filtering out potential false-negatives described in Section~\ref{ssec:false-neg_contrast}; (2) the second training step using two loss functions with only informative hard-negatives described in Section~\ref{ssec:hard-neg_contrast}. 
To calculate the similarity between the text and video moment representations projected in the IoU space, we compute the cosine similarity $s^{iou}$. To adjust the range of the final prediction, we multiply the cosine similarity by a factor of 10, following MMN. This amplification results in the final prediction $p^{iou}_{i}=\sigma(10\cdot s^{iou}_{i})$, where $\sigma$ represents the sigmoid function.
The binary cross-entropy loss is then calculated as follows:

\scriptsize
\begin{equation}
\begin{aligned}
    \mathcal{L}_{bce} = -\frac{1}{C} \sum^{C}_{i=1}{(y_{i}\log{p^{iou}_{i}}+(1-y_{i})\log{(1-p^{iou}_i)})}
\end{aligned}
\label{eq:mmn_loss1}
\end{equation}
\normalsize

\noindent where $p^{iou}_{i}$ is the confidence score of each moment, and $C$ is the total number of valid candidates.
We calculate $\mathcal{L}_{bce}$ for positive and potential true-negative samples derived in Section \ref{ssec:false-neg_contrast}.

We use the cross-directional contrastive learning loss to effectively train the model using the informative negatives derived from the methods described in Section~\ref{ssec:false-neg_contrast} and \ref{ssec:hard-neg_contrast}.
The reliable mutual matching contrastive learning loss considering the informative negative samples selection process is as follows:

\scriptsize
\begin{equation}
\begin{aligned}
    p^{*}(q_{i}|m) = \frac{\exp((f_{i}^{q\mathsf{T}}f^{m}-\psi)/\tau_{m})}{\exp((f_{i}^{q\mathsf{T}}f^{m}-\psi)/\tau_{m})+\sum_{j \in \mathcal{I}_{m}^{-}}{\exp(f_{j}^{q\mathsf{T}}f^{m}/\tau_{m})}} \:\:\:\
    \\
    p^{*}(m_{i}|q) = \frac{\exp((f_{i}^{m\mathsf{T}}f^{q}-\psi)/\tau_{q})}{\exp((f_{i}^{m\mathsf{T}}f^{q}-\psi)/\tau_{q})+\sum_{j \in \mathcal{I}_{q}^{-}}{\exp(f_{j}^{m\mathsf{T}}f^{q}/\tau_{q})}} \:\:\:\:\:\
    \\
    L_{rmm} = -(\sum^{N}_{i}{\sum^{M}_{j}\log{p^{*}(m_{ij}|q_{i})}} + \sum^{N}_{i}{\sum^{M}_{j}\log{p^{*}(q_{ij}|m_{i})}}) \:\:\:\:\:\:\:\:\:\:\:\
\end{aligned}
\label{eq:rmmn_loss1}
\end{equation}
\normalsize 

\noindent where $q_{i}$ and $m_{i}$ are corresponding positive query and moment for each moment and query, respectively.
$f^{m}$ and $f^{q}$ are moment and query features in the joint visual-text space.
$\tau_{q}$ and $\tau_{m}$ are temperatures. $\mathcal{I}^{-}$ means the indices of informative negatives derived using our sampling methods.

The final loss, denoted as $\mathcal{L}$, is formulated as a linear combination of the binary cross-entropy loss and the reliable mutual matching loss. The matching score $s$, for a specific moment given a query, is obtained by multiplying the iou score $s^{iou}$ with the reliable mutual matching score $s^{rmm}$. The reliable mutual matching score is calculated for features in the mutual matching space in the same way as the iou score.

\footnotesize
\begin{equation}
\begin{aligned}
    \mathcal{L} = \mathcal{L}_{bce} + \lambda \mathcal{L}_{rmm}, \:\:\:\:\ s = s^{iou} \cdot s^{rmm}
\end{aligned}
\label{eq:rmmn_loss3}
\end{equation}
\normalsize

%% file: file_sources/fig_method.tex
\begin{figure*}[t]
\centering
\includegraphics[width=1.0\textwidth]{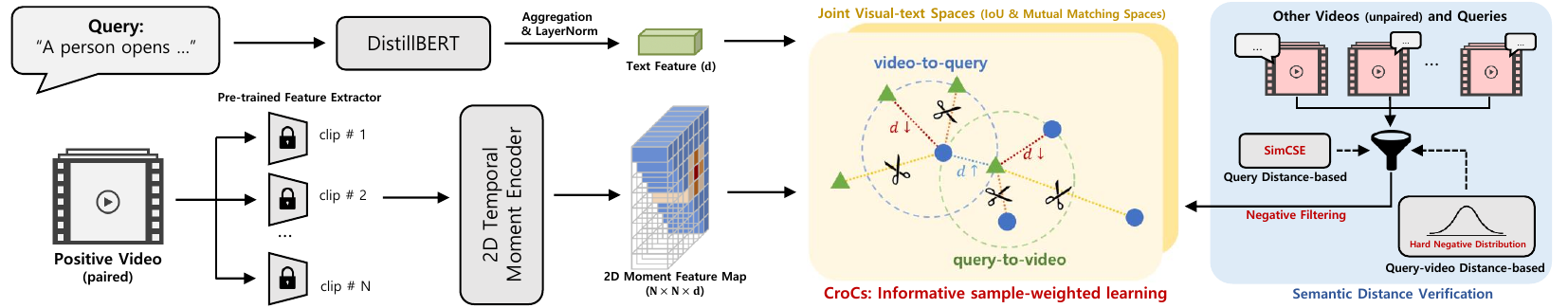}
\caption{\textbf{CroCs Overview.} We adopt the informative sample-weighted mutual matching learning to solve the \ourtask~task. The dots and triangles are the features of moments and texts. The blue dash line is matched moment-text pairs to be pulled in, while the red dash lines are negative samples of intra/inter-video to be pushed away. The yellow and orange dash lines are unmatched moment-text pairs, but not to train by filtering out since they are easy and false-negatives.}
\label{fig:method}
\end{figure*}

%% file: texts/experiments.tex
\subsection{Experimental Settings}
This section describes the experimental settings of the CroCs and other baselines. More details are shown in Appendix~\ref{sec:appendix2}.

\paragraph{\textbf{Implementation Details.}}
We use standard off-the-shelf video feature extractors without any fine-tuning: VGG \cite{simonyan2014very} feature for Charades-STA; C3D \cite{tran2015learning} feature for ActivityNet and TACoS following \cite{wang2022negative}.
We set the filtering hyper-parameter as $t_{tr}=0.9$ for the weakly-supervised potential negative contrastive learning.
We choose the number of the CroCs hard negative samples to calculate $p'(q|v)$ as 100, 200, and 100 and $p'(v|q)$ as 50, 100, and 5 for Charades-STA, ActivityNet, and TACoS, respectively.
These hyper-parameters are chosen from varying parameter setting experiments as shown in Figure \ref{fig:fig_hyper_params}.
In the hard-negative fine-tuning stage, the learning rates are degraded by multiplying 0.1.

\paragraph{\textbf{Baselines.}}
We first select the state-of-the-art models, MMN \cite{wang2022negative} and G2L \cite{li2023g2l}, as strong baselines.
As the \ourtask~ task can be formulated as an open-domain question answering (QA) task, we further choose two types of QA models that employ a cross-encoder architecture, Cross-Encoder Transformer (CET), and a dual-encoder architecture, Bi-Encoder Transformer (BET).
These architectures are widely adopted in existing \prevtask~tasks due to the closeness of the two tasks.

\vspace{-1ex}
\paragraph{\textbf{Evaluation Metrics.}}
We evaluate baselines by calculating Rank n@m.
It is defined as the percentage of queries having at least one correctly retrieved moment, i.e., IoU $\geq m$, in the top-$n$ derived moments.
The existing \prevtask~studies have reported the results of $m \in \{0.3, 0.5, 0.7\}$ with $n \in \{1, 5\}$ \cite{wang2022negative}.
But, since our \ourtask~setting searches for massive videos, we report the results of $m \in \{0.5, 0.7\}$ with $n \in \{1, 5, 20, 50\}$ for Charades-STA and AcitivityNet and $m \in \{0.5, 0.7\}$ with $n \in \{1, 5, 20\}$ for TACoS.

\subsection{\ourtask~Experimental Results}
\paragraph{\textbf{Overconfidence for the Misinformation.}}
We evaluate various VMR models (i.g., CET, G2L, and MMN) in both \prevtask~and \ourtask~settings to determine whether their performance significantly degrades when extending the search coverage to a massive video set. 
Figure~\ref{fig:fig_degrade} shows that all the baseline models undergo significant performance degradation in the \ourtask~task.
Surprisingly, although CET and G2L show quite strong performance on the \prevtask~task, their performance on the \ourtask~task notably drops.
MMN shows a relatively robust performance than CET and G2L, but it still shows significant performance drops.
They tend to be easily distracted by wrong videos and show overconfidence in the misinformation.
These outcomes indicate that the current \prevtask~task performance alone is not enough to verify the faithfulness of a \prevtask~model. 

\input{file_sources/fig_degrade}

\paragraph{\textbf{\ourtask~Performance.}}
Table~\ref{table1} shows the performance of CroCs and the baseline models on the introduced \ourtask~datasets.
We reveal that CroCs outperforms all other baselines on the \ourtask~task, and this suggests that our model correctly discriminates features of negative and positive moments.
In the Charades-STA and TACoS datasets, the models show significant improvement in performance from the weakly-supervised potential negative learning.
These results are attributed to the frequent presence of similar queries in these datasets and the possibility of false negatives.
On the other hand, the performance of ActivityNet is attributed to the cross-directional hard-negative learning since it includes various and complex text queries.

\input{file_sources/table1}

\paragraph{\textbf{\ourtask~Settings with Video Retrieval.}}
The practical usage scenario of VMR uses a pipeline integrating video retrieval with video moment retrieval.
To analyze in depth the excellence of CroCs, we also conduct \ourtask~experiments with a pipeline including a video retrieval model, PRMR model \cite{dong2022partially}.
The experimental results with video retrieval are shown in Table~\ref{table2}.
We construct a pipeline to filter top-5 logit videos from our \ourtask~retrieval pool using the PRMR model and solve the \ourtask~task.
We use the publicly deployed PRMR model on Charades-STA and ActivityNet datasets.
These experiments prove that video retrieval models are helpful to increase \ourtask~performance, and CroCs still shows superior performance than MMN in this pipeline.

\input{file_sources/table2}

%% file: file_sources/fig_degrade.tex
\begin{figure}[!h]
\newcommand\x{15}
\newcommand\p{0.99}

\begin{subfigure}[t]{\p\linewidth}
    \centering
    \includegraphics[width=8.2 cm]{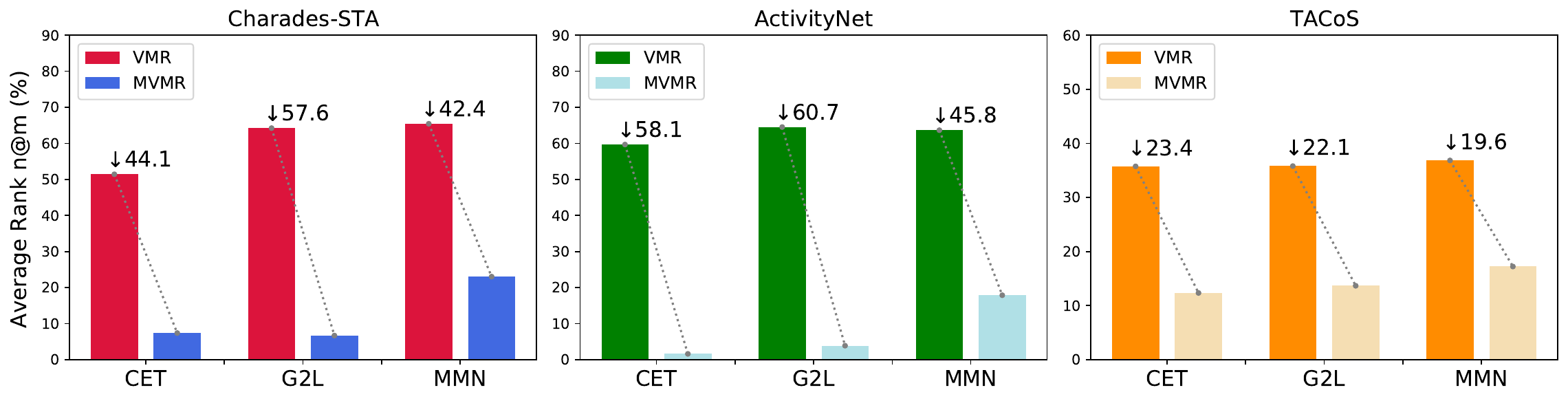}
\end{subfigure}
\vspace{-0.3cm}
\caption{\textbf{\prevtask~vs. \ourtask~Performance.} We report average scores of R1@0.5 and R5@0.5 to reveal the vulnerability of \prevtask~models to misinformation. X and Y axes correspond to the models and the average rank score, respectively.}
\vspace{-0.3cm}
\label{fig:fig_degrade}
\end{figure}

%% file: file_sources/table1.tex
\begin{table}[h]
\centering
\resizebox{1.0\linewidth}{!}
{
\begin{tabular}{@{}clcccccccc@{}}
\toprule
Dataset & \multicolumn{1}{c}{Model} & R1@0.5 & R1@0.7 & R5@0.5 & R5@0.7 & R20@0.5 & R20@0.7 & R50@0.5 & R50@0.7 \\ \midrule
\multirow{5}{*}{$\text{\ourtask}_\text{Charades-STA}$}  & CET & 2.96 & 2.02 & 11.79 & 6.97 & 32.21 & 19.91 & 50.54 & 33.88 \\
                           & BET & 2.99 & 1.94 & 10.98 & 7.00 & 25.11 & 16.55 & 37.86 & 26.08 \\
                           & G2L & 2.78 & 1.97 & 10.59 & 7.36 & 34.74 & 21.33 & 62.72 & 44.82 \\ 
                           & MMN & 12.78 & 8.40 & 33.29 & 22.36 & 57.83 & 41.71 & 73.09 & 56.75 \\ \cmidrule{2-10}
                           & CroCs$^{\dagger}$ & \underline{16.31} & \underline{10.76} & \textbf{39.05} & \underline{26.99} & \textbf{62.89} & \underline{46.39} & \underline{76.08} & \underline{60.28} \\
                           & CroCs & \textbf{17.38} & \textbf{11.44} & \underline{39.02} & \textbf{27.91} & \underline{60.98} & \textbf{46.39} & \textbf{76.29} & \textbf{60.76} \\ \midrule
                           
\multirow{5}{*}{$\text{\ourtask}_\text{ActivityNet}$} & CET & 0.90 & 0.56 & 2.32 & 1.75 & 4.73 & 3.87 & 8.02 & 6.56         \\
                            & BET & 1.16 & 0.76 & 2.55 & 2.03 & 4.64 & 3.80 & 7.10 & 5.88 \\
                           & G2L & 1.32 & 0.96 & 6.43 & 4.55 & 21.66 & 14.81 & 50.07 & 32.99 \\ 
                           & MMN & 10.71 & 6.93 & 25.04 & 16.55 & 44.50 & 30.62 & 62.09 & \underline{45.76}  \\ \cmidrule{2-10}
                           & CroCs$^{\dagger}$ & \underline{12.68} & \underline{7.87} & \underline{28.47} & \underline{18.16} & \underline{46.52} & \underline{31.12} & \underline{62.12} & 44.89 \\
                           & CroCs & \textbf{15.58} & \textbf{9.51} & \textbf{34.70} & \textbf{22.40} & \textbf{55.58} & \textbf{38.55} & \textbf{70.02} & \textbf{53.20} \\ \midrule
                           
\multirow{5}{*}{$\text{\ourtask}_\text{TACoS}$} & CET & 5.79 & 3.89 & 18.88 & 12.17 & 42.53 & 23.70 & - & - \\
                           & BET & 5.30 & 2.97 & 14.84 & 8.32 & 39.76 & 21.51 & - & - \\
                           & G2L & 5.82 & 4.31 & 21.51 & 12.29 & 42.35 & 22.52 & - & - \\ 
                           & MMN & 8.61 & 5.60 & 25.84 & 15.47 & 43.89 & 23.80 & - & - \\ \cmidrule{2-10}
                           & CroCs$^{\dagger}$ & \underline{10.75} & \underline{6.76} & \underline{27.01} & \underline{15.28} & \underline{44.38} & \textbf{25.16} & - & - \\
                           & CroCs & \textbf{10.75} & \textbf{7.30} & \textbf{29.68} & \textbf{16.50} & \textbf{44.77} & \underline{24.62} & - & - \\ \midrule
\end{tabular}
}
\caption{
\textbf{Experimental Results on \ourtask~datasets.} Bolded and under-lined results indicate the $1^{st}$ and $2^{nd}$ best performance, respectively.
CroCs$^{\dagger}$ is the model using only weakly-supervised potential negative learning (\S \ref{ssec:false-neg_contrast}).
CroCs uses both of two informative sample-weighted learning methods (\S \ref{ssec:false-neg_contrast} and \S\ref{ssec:hard-neg_contrast}).
We train all the baselines by three trials and report the averaged \ourtask~scores.
}
\vspace{-0.5cm}
\label{table1}
\end{table}

%% file: file_sources/table2.tex
\begin{table}[h]
\small
\centering
\resizebox{1.0\linewidth}{!}{\begin{tabular}{c c c c c c c c}
\hline
\textbf{Dataset} & \textbf{Model} & \textbf{R1@0.3} & \textbf{R1@0.5} & \textbf{R5@0.3} & \textbf{R5@0.5} \\ 
\hline
\multirow{2}{*}{\textbf{$\text{\ourtask}_\text{Charades-STA}$}} & \textbf{MMN} & 32.15 & 27.10 & 65.97 & 57.66 \\
& \textbf{CroCs} & \textbf{36.64} & \textbf{29.27} & \textbf{68.06} & \textbf{59.78} \\
\hline
\multirow{2}{*}{\textbf{$\text{\ourtask}_\text{ActivityNet}$}} & \textbf{MMN} & 32.39 & 25.04 & 66.84 & 55.03 \\
& \textbf{CroCs} & \textbf{39.25} & \textbf{29.59} & \textbf{71.54} & \textbf{60.02} \\
\hline
\end{tabular}}
\caption{\textbf{\ourtask~Experiments with Video Retrieval.}}
\vspace{-0.3cm}
\label{table2}
\end{table}


%% file: texts/conclusion.tex
In this paper, we propose the Massive Videos Moment Retrieval for Faithfulness Evaluation (\ourtask) task, which aims to detect a moment for a natural language query from a massive video set to evaluate the faithfulness of video moment retrieval.
To stimulate the research, we propose a simple automatic method to construct trustful \ourtask~datasets by employing semantic distance verification methods, and finally build three \ourtask~datasets by extending existing \prevtask~datasets.
We further introduce a robust informative sample-weighted learning method called CroCs to distinguish positives from negative distractors effectively in solving the \ourtask~task.

%% file: texts/ack.tex
\begin{acks}
This work was supported by Adobe Research and LG AI Research.
This work was partly supported by Samsung Electronics.
This work was partly supported by Institute of Information \& communications Technology Planning \& Evaluation (IITP) grant funded by the Korea government (MSIT) [No.RS-2022-II220184, Development and Study of AI Technologies to Inexpensively Conform to Evolving Policy on Ethics \& No.RS-2021-II211343, Artificial Intelligence Graduate School Program (Seoul National University) \& No.RS-2021-II212068, Artificial Intelligence Innovation Hub (Artificial Intelligence Institute, Seoul National University)].
K. Jung is with ASRI, Seoul National University, Korea.
The Institute of Engineering Research at Seoul National University provided research facilities for this work.
\end{acks}

%% file: texts/appendix.tex
\section{Analysis on Filter Methods}
\label{sec:appendix_analysis_filters}
We construct trustful \ourtask~datasets using SimCSE and EMScore. The visualizations of the results derived by SimCSE and EMScore for the TACoS dataset are shown in Figure~\ref{fig:dist_tacos}-(a).
We select \textit{"The person peels a kiwi"} to show the distribution of SimCSE embeddings.

\input{file_sources/fig_dist_tacos}

Figure~\ref{fig:semeval} shows the distribution of SimCSE score and human-annotated STS16\cite{agirre2016semeval} dataset score.
STS16 dataset provides text pairs and annotated similarity scores (five-point Likert scale).
In the dataset, two texts are regarded as similar when the score is higher than 3.
We visualize the distribution of SimCSE and human annotated scores for all text pairs in the dataset, and select $t^{+}_{te}=0.9$ and $t^{-}_{te}=0.5$ as thresholds for the SimCSE filtering to derive MVMR positive and negative candidates, respectively.
STS16 dataset includes text pairs that are challenging to distinguish. However, the three datasets utilized in our experiment do not demand similarity measurement at such a complex level. Consequently, we can reliably discern the similarity between queries using the provided thresholds.

\input{file_sources/fig_semeval.tex}

\section{Implementation Details}
\label{sec:appendix2}
We use standard off-the-shelf video feature extractors without any fine-tuning.
We use VGG \cite{simonyan2014very} feature for Charades-STA and C3D \cite{tran2015learning} feature for ActivityNet and TACoS following the previous work \cite{wang2022negative}.
We train all VMR models on 1 NVIDIA RTX A6000 GPU and early stop by averaging scores of all evaluation metrics.
 
\subsection{CroCs}
\label{sec:appendix-rmmn}
We implement CroCs by following MMN's model architecture.
CroCs represents each query and video moment using query and video encoders to a joint visual-text space.
From the joint visual-text space, CroCs computes the similarity between the query and video moment representations and selects the best-matched video moment.

\paragraph{\textbf{Query and Video Encoders.}} We choose DistilBERT \citep{sanh2019distilbert} for the text query encoder following MMN since it is a lightweight model showing significant performance.
We calculate the representation of each query $f^{q} \in \mathbb{R}^{d}$ using the global average pooling over all tokens.
We adopt the approach of encoding the input video as a 2D temporal moment feature map, inspired by 2D-TAN \cite{zhang2020learning} and MMN.
To achieve this, we segment the input video into video clips denoted as $\{c_{i}\}_{i=1}^{l_{c}/k}$, where each clip $c_{i}$ consists of $k$ frames. $l_{c}$ means the total number of frames in the video.
The clip-level representations are extracted using a pre-trained visual model (e.g., C3D).
By sampling a fixed length with a stride of $\frac{l_{c}}{k\cdot N}$, we obtain $N$ clip-level features.
These features are then passed through an FC layer to reduce the dimensionality, resulting in the features $\{f^{v}_{i}\}^{N}_{i=1}$, where $f^{v}_{i} \in \mathbb{R}^{d}$.
Utilizing these features, we construct a 2D temporal moment feature map $F \in \mathbb{R}^{N \times N \times d}$ by employing max-pooling as the moment-level feature aggregation method following the baseline \citep{wang2022negative}.
Additionally, we generate a 2D feature map $F'$ of the same size by passing $F$ to 2D Convolution, allowing the representation of moment relations as employed in MMN.

\paragraph{\textbf{Joint Visual-Text Space.}}
Initially, we apply layer normalization to both the video moments and text features. Subsequently, we utilize a linear projection layer and a 1$\times$1 convolution layer to project the text and video in the same embedding space, respectively. The projected features are then employed in two distinct representational spaces: the IoU space and the mutual matching space. These spaces serve as the basis for computing the binary cross-entropy loss and the contrastive learning loss, respectively.

\paragraph{\textbf{Hyperparameter Settings.}} Our convolution network for deriving 2D features is exactly the same as MMN, including the number of sampled clips $N$, the number of 2D conv layers $L$, kernel size, and channels.
We set the dimension of the joint feature visual-text space $d=256$, and temperatures $\tau_{m}=\tau_{q}=0.1$. We utilize the pre-trained HuggingFace implementation of DistilBERT \cite{wolf2019huggingface}. We set margin $\psi$ as 0.4, 0.3, and 0.1 for Charades-STA, ActivityNet, and TACoS, respectively.
We use AdamW \cite{loshchilov2017decoupled} optimizer with learning rate of $1 \times 10^{-4}$, $8 \times 10^{-4}$, $1.5 \times 10^{-3}$ for Charades-STA, ActivityNet, and TACoS, respectively.
We set $\lambda$ as 0.05 for Charades-STA and TACoS and 0.1 for ActivityNet.
We set the reliable true-negatives filtering threshold $t_{tr}=0.9$.
We set the number of hard-negative samples to calculate $p'(q|v)$ as 100, 200, and 100 and $p'(v|q)$ as 50, 100, and 5 for Charades-STA, ActivityNet, and TACoS, respectively.
The experimental results for various hyper-parameters for the hard-negative sampling are shown in Figure~\ref{fig:fig_hyper_params}.

\input{file_sources/fig_hyperparams}

\subsection{Baselines}
\noindent\textbf{MMN} \cite{wang2022negative} adopts a dual-modal encoding design to get video clip and text representations.
We utilized the publicly deployed code\footnote{https://github.com/MCG-NJU/MMN} to implement our own MMN model for our experiment using exactly the same hyper-parameters setting.

\noindent\textbf{G2L} \cite{li2023g2l} is a sort of stat-of-the-art model, extending MMN architecture.
We utilized the publicly deployed code\footnote{https://github.com/lihxxxxx/G2L} to implement our own G2L model for our experiment using exactly the same hyper-parameters setting.

\noindent\textbf{Cross-Encoder Transformer (CET)} is a cross-modal encoding model implemented based on a 6-layered transformer architecture.
It is designed to predict the start and end points of a video moment by utilizing a model architecture, which is used in the QA task.
We concatenate the pre-extracted video clip features and the text features derived from the Distil-BERT to use as an input of the transformer.
To concatenate two extracted features, we should equalize the dimensions of two features with a linear layer as $\bar{H}^{v} = H^{v}W^{v}$ and $\bar{H}^{q} = H^{q}W^{q}$, where $H^{v} \in \mathbb{R}^{l_{v}, d^{v}}$, $H^{q} \in \mathbb{R}^{l_{q}, d^{q}}$, $W^{v} \in \mathbb{R}^{d^{v}, d}$, $W^{q} \in \mathbb{R}^{d^{q}, d}$.
$H^{v}$ is the pre-extracted video clip features and $H^{q}$ is the features extracted using Distil-BERT.
And then, we derive the representation of each video clip feature from the last layer of the transformer-encoder as follows:

\footnotesize
\begin{equation}
\begin{aligned}
    H^{v,q} = TransformerEnc(\bar{H}^{v} \oplus \bar{H}^{q})
\end{aligned}
\label{eq:apx2}
\end{equation}
\normalsize

where $H^{v,q} \in \mathbb{R}^{l_{vq}, d}$, $l_{vq}=l_{v}+l_{q}$. 
We utilize only the representation part of video features, and the 3-layered MLPs and Sigmoid follow them to predict the start and end positions of the moment, respectively, as follows:

\footnotesize
\begin{equation}
\begin{aligned}
    p^{s} = Sigmoid(MLP_{s}(H^{v,q}_{1:l_{v}})), \:\:\:\ p^{e} = Sigmoid(MLP_{e}(H^{v,q}_{1:l_{v}}))
\end{aligned}
\label{eq:apx3}
\end{equation}
\normalsize

where $p^{s}$ and $p^{e}  \in \mathbb{R}^{l_{v}}$.
We also introduce the interpolation labels $y_{i}^{s}$, $y_{i}^{e}$ to utilize abundant information of moment labels as follows:

\footnotesize
\begin{equation}
\begin{aligned}
y_{i}^{s} = \left\{ 
  \begin{array}{ c l }
    ({\frac{x_{e}-i}{x_{e}^{*}-x_{s}^{*}+1}})^{k} & \quad \textrm{if } x_{s}^{*} \leq i \leq x_{e}^{*} \\
    0                 & \quad \textrm{otherwise}
  \end{array}
\right. \:\:\:\:\:\:\\
y_{i}^{e} = \left\{ 
  \begin{array}{ c l }
    ({\frac{i-x_{s}}{x_{e}^{*}-x_{s}^{*}+1}})^{k} & \quad \textrm{if } x_{s}^{*} \leq i \leq x_{e}^{*} \\
    0                 & \quad \textrm{otherwise}
  \end{array}
\right. \:\:\:\:\:\
\end{aligned}
\label{eq:apx4}
\end{equation}
\normalsize

We finally use the binary cross-entropy to calculate losses for the pair of $(p_{i}^{s}, y_{i}^{s})$ and $(p_{i}^{e}, y_{i}^{e})$ as $\mathcal{L}^{s} = BCE(p^{s}, y^{s})$ and $\mathcal{L}^{e} = BCE(p^{e}, y^{e})$, respectively. BCE means the binary cross-entropy.
The final loss to train the CET model is defined as follows: 

\footnotesize
\begin{equation}
\begin{aligned}
\mathcal{L} = \frac{\mathcal{L}^{s} + \mathcal{L}^{e}}{2}
\end{aligned}
\label{eq:apx6}
\end{equation}
\normalsize

We use $k=10$ and $d=512$ as a hyperparameter and AdamW optimizer with learning rate 1e-3.

\noindent\textbf{Bi-Encoder Transformer (BET)} is a dual-modal encoding model that encodes text and video clip features independently.
The pre-extracted video clip features are encoded using a 6-layered transformer, and text features are encoded using the Distil-BERT, followed by the average-pooling function and 3-layered MLPs.

\footnotesize
\begin{equation}
\begin{aligned}
    \tilde{H}^{v} = TransformerEnc(\bar{H}^{v})
\end{aligned}
\label{eq:apx7}
\end{equation}
\normalsize

\footnotesize
\begin{equation}
\begin{aligned}
    \tilde{H}^{s} = MLP_{s}(Pool(\bar{H}^{q})), \:\:\:\ \tilde{H}^{e} = MLP_{e}(Pool(\bar{H}^{q}))  \\
\end{aligned}
\label{eq:apx8}
\end{equation}
\normalsize

\noindent where $\tilde{H}^{v} \in \mathbb{R}^{d}$, $\tilde{H}^{s} \in \mathbb{R}^{d}$ and $\tilde{H}^{e} \in \mathbb{R}^{d}$.
$Pool$ means the average pooling.
We calculate the cosine similarity between the encoded video clip representations and the encoded start and end representations to predict a moment.

\footnotesize
\begin{equation}
\begin{aligned}
    p^{s} = Sigmoid(r*cosine(\tilde{H}^{v}, \tilde{H}^{s})),\:\:\:\  p^{e} = Sigmoid(r*cosine(\tilde{H}^{v}, \tilde{H}^{e}))  \\
\end{aligned}
\label{eq:apx9}
\end{equation}
\normalsize

We use the interpolation labels and the binary cross-entropy loss for the BET, similar to CET.
We use $r=10$ and $d=512$ as a hyperparameter and AdamW optimizer with learning rate 1e-3.



%% file: file_sources/fig_dist_tacos.tex
\begin{figure}[!h]
\centering
\includegraphics[width=6cm]{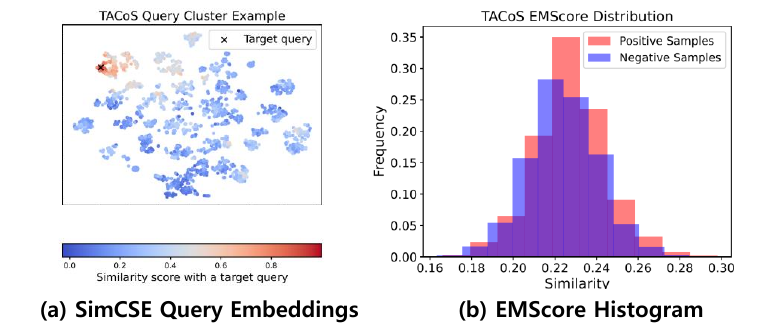}
\vspace{-.3cm}
\caption{\textbf{Qualitative Analysis for Filtering Methods for the TACoS Dataset.} }
\vspace{-.3cm}
\label{fig:dist_tacos}
\end{figure}

%% file: file_sources/fig_semeval.tex
\begin{figure}[!h]
\vspace{-.3cm}
\centering
\includegraphics[width=5.5cm]{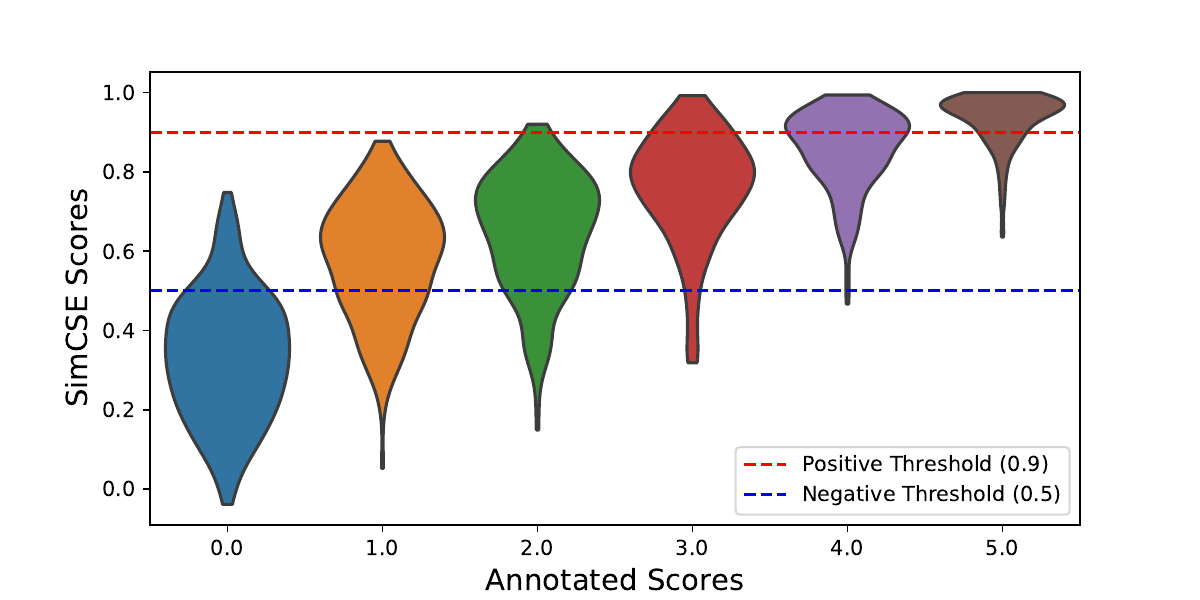}
\vspace{-0.3cm}
\caption{\textbf{The distribution of SimCSE and Human-annotated Scores.} The X and Y axes mean the human-annotated and the SimCSE scores, respectively. The width of each graph corresponds to the number of samples.}
\vspace{-.3cm}
\label{fig:semeval}
\end{figure}

%% file: file_sources/fig_hyperparams.tex
\begin{figure}[!h]
\newcommand\x{15}
\newcommand\p{0.99}

\begin{subfigure}[t]{\p\linewidth}
    \centering
    \includegraphics[width=8.2 cm]{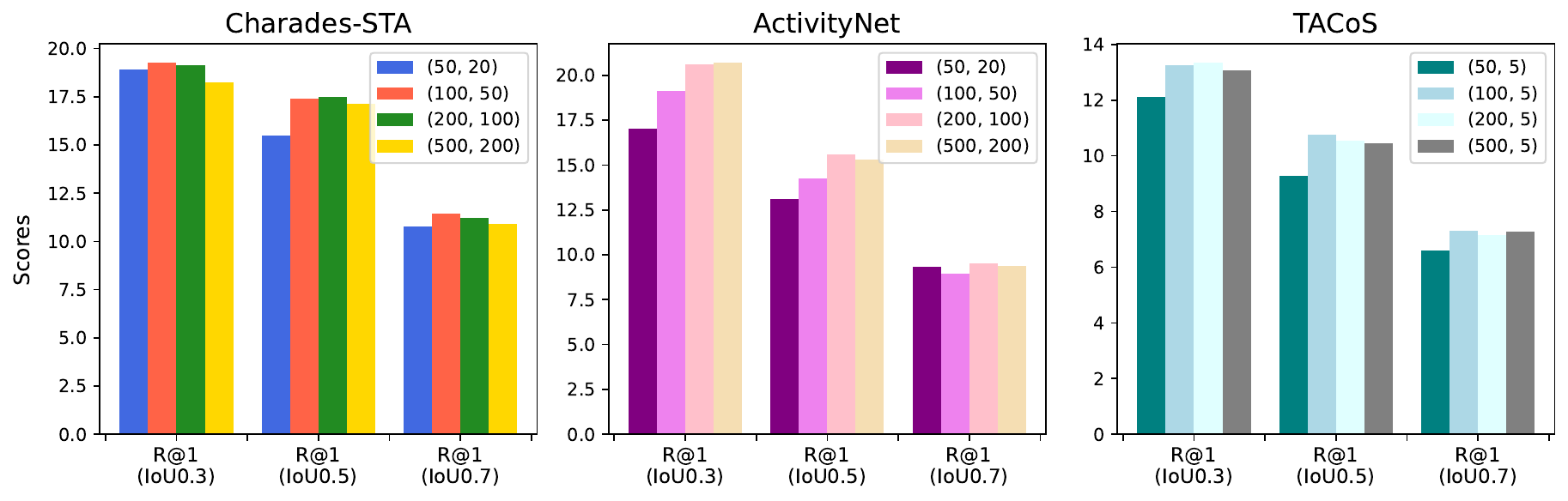}
\end{subfigure}

\caption{\textbf{Evaluation results for various hyper-parameters.} We measure Rank 1 (IoU>m), where m$\in\{0.3, 0.5, 0.7\}$ to find the best hyper-parameters. Each $(x, y)$ means the number of negative samples. $x$ and $y$ mean a number of negative samples for $p'(v|q)$ (query-to-video) and $p'(q|v)$ (query-to-video), respectively. We select the number of hard negative samples (\S\ref{ssec:hard-neg_contrast}) to calculate $p'(q|v)$ as 100, 200, and 100 and $p'(v|q)$ as 50, 100, and 5 for Charades-STA, ActivityNet, and TACoS, respectively.}
\label{fig:fig_hyper_params}
\end{figure}